\documentclass[sigconf]{acmart}

\AtBeginDocument{%
  \providecommand\BibTeX{{%
    \normalfont B\kern-0.5em{\scshape i\kern-0.25em b}\kern-0.8em\TeX}}}

\copyrightyear{2021}
\acmYear{2021}
\setcopyright{acmlicensed}
\acmConference[MM '21]{Proceedings of the 29th ACM International Conference on Multimedia}{October 20--24, 2021}{Virtual Event, China}
\acmBooktitle{Proceedings of the 29th ACM International Conference on Multimedia (MM '21), October 20--24, 2021, Virtual Event, China} \acmPrice{15.00}
\acmDOI{10.1145/3474085.3475210} 
\acmISBN{978-1-4503-8651-7/21/10}

\begin{document}
\fancyhead{}
\title{Shape Controllable Virtual Try-on for Underwear Models}


\author{Xin Gao$^1$,Zhenjiang Liu$^1$,Zunlei Feng$^2$,Chengji Shen$^2$,Kairi Ou$^1$,Haihong Tang$^1$,Mingli Song$^2$}
\affiliation{
\institution{$^1$Alibaba Group,$^2$Zhejiang University}
 }
 \email{
{zimu.gx, stan.lzj, piaoxue}@alibaba-inc.com,{zunleifeng, chengji.shen, brooksong}@zju.edu.cn,
{suzhe.okr}@taobao.com}
 \authornote{Corresponding authors}
\renewcommand{\shortauthors}{Gao and Liu, et al.}

\begin{abstract}
Image virtual try-on task has abundant applications and has become a hot research topic recently. Existing 2D image-based virtual try-on methods aim to transfer a target clothing image onto a reference person, which has two main disadvantages: cannot control the size and length precisely; unable to accurately estimate the user's figure in the case of users wearing thick clothes, resulting in inaccurate dressing effect.
In this paper, we put forward an akin task that aims to dress clothing for underwear models.
To solve the above drawbacks, we propose a Shape Controllable Virtual Try-On Network (SC-VTON), where a graph attention network integrates the information of model and clothing to generate the warped clothing image. 
In addition, the control points are incorporated into SC-VTON for the desired clothing shape. 
Furthermore, by adding a Splitting Network and a Synthesis Network, we can use clothing/model pair data to help optimize the deformation module and generalize the task to the typical virtual try-on task. 
Extensive experiments show that the proposed method can achieve accurate shape control. Meanwhile, compared with other methods, our method can generate high-resolution results with detailed textures.
\end{abstract}

\ccsdesc[500]{Applied computing~Online shopping}
\ccsdesc[500]{Computing methodologies~Image-based rendering}

\keywords{Virtual try-on; Graph attention networks; Image warping}


\maketitle

\section{Introduction}\label{section_1}

\begin{figure}
    \centering
    \includegraphics[width=0.49\textwidth]{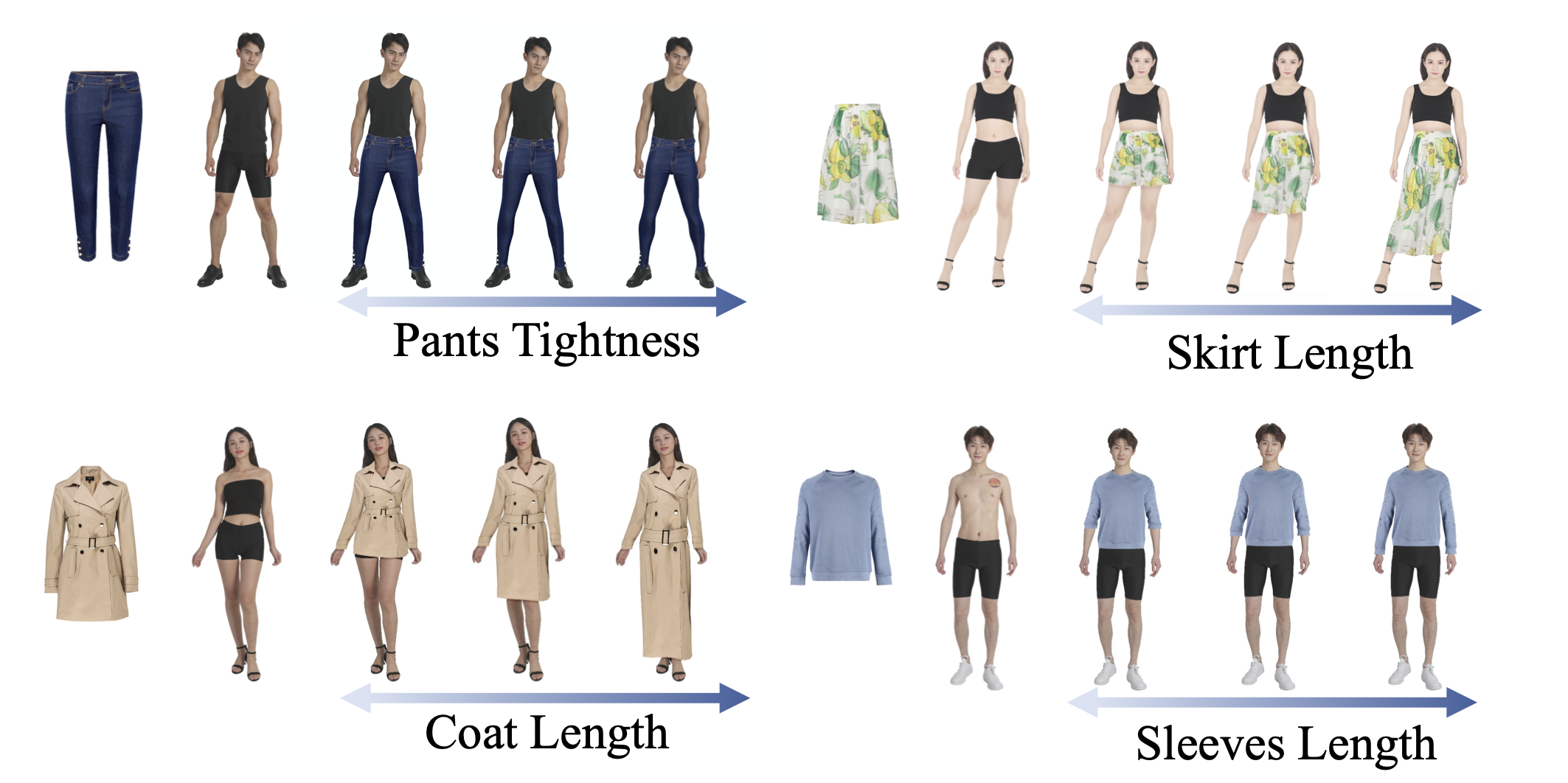}
    \caption{Shape controllable visual results.
    The shape can be continuously controlled in length and tightness.
    }
    \label{fig:1}
\end{figure}

The online virtual try-on system \cite{han2019clothflow,han2018viton,issenhuth2019end,issenhuth2020not,jandial2020sievenet,li2020toward,Minar_C3DVTON_2020_CVPR_Workshops,raffiee2020garmentgan,ren2020deep,roy2020lgvton,wang2018toward,wu2019m2e,yang2020towards,yu2019vtnfp} has become a research hot-spot in recent years to accommodate the vast market demand.
Virtual try-on applications provide users with convenient what-you-see-is-what-you-get function: users can pick a clothing and put it on themselves virtually, and matching their favorite. 
It is still a big challenge for virtual try-on algorithms to make the results both realistic and attractive.

Existing virtual try-on techniques can be divided into 3D virtual try-on methods and 2D image-based methods. 3D virtual try-on methods need to model characters and clothing, then render the digitalized clothing on the character, achieving controllable and detailed results. 
However, the process of modeling for characters and clothing is costly and time-consuming.
Rendering process also requires high performance of mobile devices.

Han et al. \cite{han2018viton} proposed a 2D image-based virtual try-on task that adopts model and clothing pairs to train a clothing deformation sub-network and then composites the deformed clothing and coarse model with a generator.
Following \cite{han2018viton}, many 2D image-based methods \cite{issenhuth2020not,wang2018toward,yang2020towards,yu2019vtnfp} have been proposed. 
Most of them focus on finding a better deformation module or a more realistic generator. 
However, few of those methods could control clothing shape, which is a significant capacity for the virtual try-on task. 
What's more, existing 2D virtual try-on methods use parsing and key points information of the clothed model image as input,
which cannot accurately obtain the model's figure and leads to inaccurate dressing effects. 
Limited by low-resolution datasets and methodological limitations, most methods can only generate low-resolution results and have obvious blurry visual results on high-resolution results, which is not applicable in the online e-commerce scene.

This paper proposes the task that dressing clothing for underwear models, which is also an urgent demand for online clothing shops to exhibit new clothing efficiently.
The model's figure can be more easily got when they wear underwear than thick clothing.

Considering the real application requirement and the drawbacks of existing 2D image-based methods, we propose a Shape Controllable Virtual Try-On Network (SC-VTON) for the underwear model dressing task. As shown in Fig. \ref{fig:1}, SC-VTON can achieve continuous shape control in tightness and length. 
SC-VTON uses GAT (Graph Attention Network) \cite{velivckovic2017graph} to learn the deformation offset of the meshed clothing. A shape information extraction module is used to extracts the features of the underwear model and clothing. The extracted feature maps are concatenated as shape vectors in channel-dimension. With the guidance of shape vectors and control points (the control points are some discrete points near the shape of the model), GAT predicts coordinate offsets of all mesh vertices. The warped clothing image is obtained by using the differentiable rendering module. The underwear model and clothing pairs for training SC-VTON are generated with an annotation tool based on As Rigid As Possible (ARAP) \cite{sorkine2007rigid} with clothing key points as control points.

Furthermore, in-shop clothing/model pairs are adopted to optimize SC-VTON. To generate self-loop supervised information, we devise a splitting network to predict the underwear model, and a synthesis network to synthesize the underwear model with the clothing image. The self-loop supervised information will enhance the robustness of SC-VTON in the real application by combining SC-VTON with the pre-trained splitting network and synthesis network. What's more, with the splitting and synthesis networks, the proposed SC-VTON can be easily extended to normal virtual try-on for a model in clothing.

Therefore, our approach is the first method for the underwear model dressing task, which can adaptively generate high-quality results with different resolutions. 
The proposed SC-VTON is the first method that achieves continuous shape control in length and tightness.
What's more, the devised self-loop optimization improves the robustness of SC-VTON and extends the framework to the typical virtual try-on task for a model in clothing. Extensive experiments demonstrate that our method achieves more pleasing visual results than the state-of-art methods.

\section{Related Works}\label{section_2}

\subsection{Virtual Try-on}\label{section_2_1}
Virtual try-on systems have been an attractive topic even before the renaissance of deep learning. 

3D virtual try-on systems based on human reconstruction in computer graphics have been widely researched, such as \cite{alldieck2019learning,bhatnagar2019multi,li2010fitting,loper2015smpl,umetani2011sensitive}. While Physics-Based Simulation (PBS) can accurately drape a 3D garment on a 3D body \cite{gundogdu2019garnet,patel2020tailornet,santesteban2019learning,vidaurre2020fully}, it remains too costly for real-time applications. 
3D human estimation methods try to digitalize real-world 2D character photos to 3D models, which have made significant progress after the popularity of deep learning, such as \cite{guler2019holopose,kanazawa2018end,kolotouros2019convolutional,xu2019denserac}, while \cite{habermann2020deepcap,jiang2020bcnet,saito2019pifu,saito2020pifuhd} also digitizing with garments, these methods still need to be optimized in texture authenticity.

Recently 2D image-based virtual try-on tasks have gained much attention from academia and industry. It is highly challenging because it requires to warp the clothing on the target person while preserving its patterns and characteristics. 

Han et al. \cite{han2018viton} proposed a 2D image-based virtual try-on task that adopts model and clothing pairs to train a clothing deformation sub-network and then composites the deformed clothing and coarse model with a generation network.
Based on \cite{han2018viton}, Wang et al. \cite{wang2018toward} refined the architecture with a geometric matching module to learn parameters for Thin-Plate Splines (TPS) deformation \cite{wood2003thin}.
Considering the complexity of pose and self-occlusion cases, Yu et al. \cite{yu2019vtnfp} and Raffiee et al. \cite{raffiee2020garmentgan} took parsing information into account and used a generator to predict the target model's parsing result, which could generate better results in self-occlusion cases.
What's more, Wu et al. \cite{wu2019m2e} adopted DensePose \cite{alp2018densepose} to get more precise information of the model. Minar et al.  \cite{Minar_C3DVTON_2020_CVPR_Workshops} solved the problem by first mapping 2D clothing texture to 3D character model and deforming the clothing in 3D space, and then back to 2D space.  Chaudhuri et al. \cite{chaudhuri2021semi} proposed a method to generate diverse high fidelity texture maps for 3D human meshes in a semi-supervised setup. 
On the other hand, some researchers focus on the improvement of the deformation module. In \cite{issenhuth2019end,issenhuth2020not,jandial2020sievenet,yang2020towards}, the spatial transformer networks \cite{jaderberg2015spatial} is adopted to get more accurate deformation result. Roy et al. \cite{roy2020lgvton} adopted human and fashion landmarks to help optimize thin-plate splines deformation model. Han et al. \cite{han2019clothflow} and Ren et al. \cite{ren2020deep} adopted optical flow to get a more natural texture deformation effect. Li et al.  \cite{li2020toward} tried to get better results by using multiple specialized warps. Ge et al. \cite{ge2021parser} proposed a knowledge distillation method to produce images without human parsing. 

Based on the generated high-resolution images of StyleGAN \cite{karras2019style},
Yildirim et al. \cite{yildirim2019generating} trained a conditional StyleGAN for generating virtual try-on images with clothing image as conditional input on around 380K entries. Cheng et al.  \cite{cheng2020adgan} embedded human geometries into the latent space as independent codes and achieved flexible and continuous control of geometries via mixing and interpolation operations in explicit style representations. 
However, acceptable visual results of StyleGAN usually requires hundreds of thousands of samples, which is very difficult to obtain. 

\begin{figure*}
    \centering
    \includegraphics[width=1.0\textwidth]{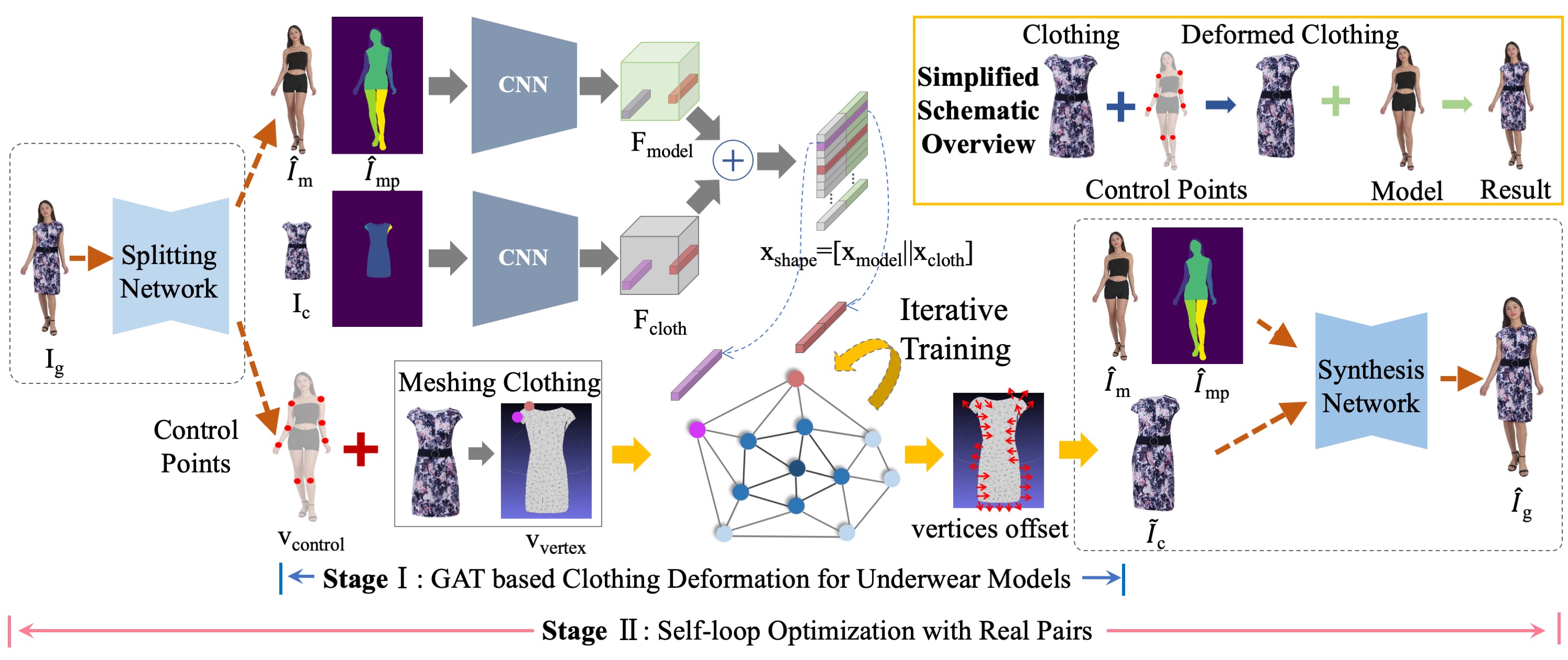}
    \caption{
    An overview of the whole framework. Stage \uppercase\expandafter{\romannumeral1} : The shape information extraction model extract feature maps from the underwear model and clothing image with CNN extractors. The clothing is meshed and presented as a graph. GAT based deformation module predicts the coordinate offsets with the clothing graph as input under the guidance of control points and shape information extracted from the model and clothing feature maps. The warped clothing image is obtained by using the differentiable rendering module. Stage \uppercase\expandafter{\romannumeral2} : The splitting network and  the synthesis network are added into SC-VTON (Stage \uppercase\expandafter{\romannumeral1}). The whole framework is extended to the typical virtual try-on task for clothed model image. Then, the whole framework can be trained with real pairs (clothing image and clothed model image).}
    \label{fig:2}
    \vspace{-0.8em}
\end{figure*}

\subsection{GCN based Deformation}\label{section_2_2}
Graph Convolutional Network (GCN) \cite{kipf2016semi} extends convolution operation from euclidean data (such as images) to a more extensive topological structure and is widely used in natural language processing, recommendation system, computer vision, and computer graphics domains. 
Wang et al.  \cite{wang2018pixel2mesh} represented 3D meshes in a graph-based CNN and produced correct geometry by progressively deforming an ellipsoid. Based on \cite{wang2018pixel2mesh}, Wen et al. \cite{wen2019pixel2mesh++} used GCN to solve the problem of shape generation in 3D mesh representation from a few color images with known camera poses.  
For the virtual try-on task, Vidaurre \cite{vidaurre2020fully} used GCN for parametric predefined 2D panels with arbitrary mesh topology. 
Kolotouros et al. \cite{kolotouros2019convolutional} adopted a Graph-CNN to estimate 3D human pose and shape from a single image.
Zhu et al. \cite{zhu2020deep} used GCN to predict 3D clothing's feature line and use it to warp clothing.

Graph Attention Networks (GATs) \cite{velivckovic2017graph} enable specifying different weights to
different nodes in a neighborhood, leveraging masked self-attentional layers to
address the shortcomings of prior methods based on graph convolutions. In this work, we consider using GAT to get more precise warping results.

\section{Methods}\label{section_3}

Different from the typical virtual try-on task, the primary goal of the paper is to wear clothing for an underwear model: given an underwear model image $I_m$ and a clothing image $I_c$, try to generate model image $I_g$ wearing the clothing with the same pose as $I_m$.
It is hard to get the pair $\{I_m, I_c, I_g\}$ that the underwear models and the model wearing clothing with the same pose at the same time.
We adopt an annotation tool based on ARAP with manually fine-tuning to get the pseudo-labeled pairs $\{I_m, I_c, I_{g'}\}$, which are used as ground truth pairs for SC-VTON on stage \uppercase\expandafter{\romannumeral1} (Section \ref{section_3_1}).
Furthermore, real pairs $\{I_c, I_g\}$ are used to improve the robustness of SC-VTON in application scenarios. 
To adapt to real pairs $\{I_c, I_g\}$, a splitting network and a synthesis network are added into SC-VTON, which is extended to handle virtual try-on task for model in clothing (Section \ref{section_3_2}). The whole framework is illustrated in Fig. \ref{fig:2}.

\subsection{GAT based Clothing Deformation for Underwear Models}\label{section_3_1}
The GAT based clothing deformation contains three parts: the shape information extraction module, the control points guided deformation module and the differentiable rendering module. 

The shape information extraction module is devised for extracting features from the clothing image and the underwear model image.
The extracted feature maps will be concatenated as shape vectors and sent to the deformation module as part of the guiding information. 
For the control points guided deformation module, the clothing firstly meshes with the Delaunay Triangulation Algorithm\cite{lee1980two}, then the meshed coordinate points are set as vertices $V$ of a graph $G=(V,E)$, where $E$ denotes the edge set (See \emph{supplementary materials} for more details). For each vertex in $V$, the vertex's information is expressed as a vertex vector.
Next, the shape control points and the shape vectors are integrated into the corresponding vertex vector as the deformation's guiding information.
GAT predicts coordinate offsets of all mesh vertices. The warped clothing image is obtained by using the differentiable rendering module. 
The whole deformation process is trained in an iterative manner to obtain higher accuracy. 
The following subsections give detailed descriptions.

\subsubsection{Shape Information Extraction Module}\label{section_3_1_1}
The shape information extraction module contains two branches. 
The first branch extracts model's shape feature maps $F_{model}$ from the original model image and model's parsing image with a CNN extractor.
The second branch extracts clothing feature maps $F_{cloth}$ from the original clothing image and clothing's parsing image with the same architecture (not sharing weights).
It is worth noting that the clothing image has been coarsely aligned to the model image with some key points as references.
For each vertex in the clothing graph, the Perceptual Feature Pooling \cite{wang2018pixel2mesh} is adopted to find the mapping relationship between each vertex's coordinate and the feature vector's coordinate in the extracted feature maps $F_{model}$ and $F_{cloth}$. Next, feature vectors $x_{model} \in F_{model}$ and $x_{cloth} \in F_{cloth}$ with the same coordinate are concatenated as shape vectors $x_{shape}=[x_{model} || x_{cloth}]$, which will be appended to the vertex vector $x$ (here || denotes concatenation). 

\subsubsection{Control Points Guided Deformation Module}\label{section_3_1_2}
Here we first introduce how to get the meshed clothing. A Deeplabv3+ model \cite{chen2018encoder} is trained to get clothing's parsing results. Then, we uniformly sample points along the contour of clothing according to the parsing results.
Next, the Delaunay Triangulation \cite{lee1980two} is adopted to mesh the inner part of the clothing. Contour points and inner points form the vertex set $V$ (See \emph{supplementary materials} for more details).
Each vertex's information is expressed as a vertex vector $x = [v_{vertex} || v_{control} || x_{shape}]$, which is composed of three parts: the vertex's coordinate $v_{vertex}$, the control point's coordinate $v_{control}$ and the corresponding shape feature vector $x_{shape}$.

\begin{figure}[!t]
    \centering
    \includegraphics[width=0.49\textwidth]{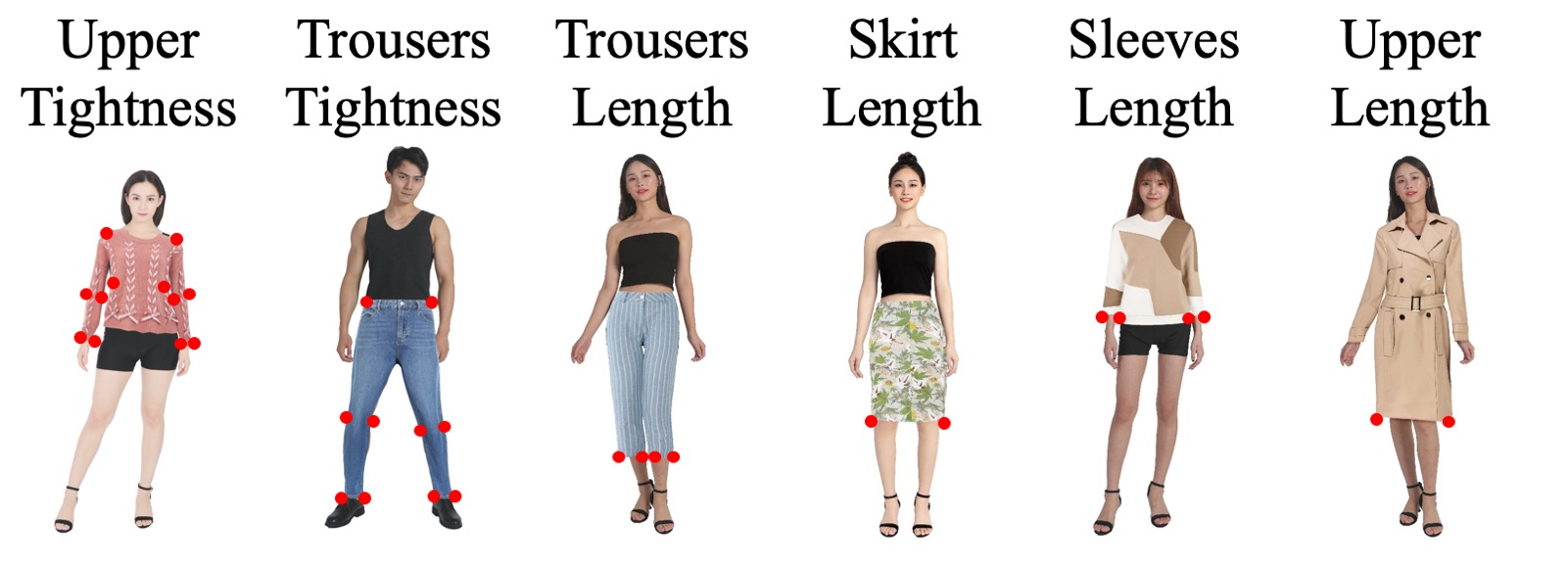}
    \caption{Control points definitions. The control points are discrete points near the silhouette of the model. The red points show the control points with different clothing types.}
    \label{fig:3}
\end{figure}

Control points are some discrete points near the silhouette of the model. The red points in Fig. \ref{fig:3} show control points with different clothing types. Control points are obtained indirectly through model key points and model parsing results. SC-VTON can control both the tightness and length of the clothing, points near the model's limbs and torso are used. By slightly adjusting the control points' position (For length, adjusting the points position along the extension of the limbs. For tightness, a dilate operation is used for the model's parsing result and finding the corresponding position according to the dilated parsing), we can achieve a shape controllable virtual try-on result.

GAT introduces the attention mechanism as a substitute for the statically normalized convolution operation. Below are the equations to compute the node embedding $h_{i}^{l+1}$ of layer $l+1$ from the embeddings of layer $l$.
\begin{equation}
\begin{aligned}
& z_{i}^{l} = W^{l} h_{i}^{l}, \\
& \alpha_{ij}^{l} = \frac{ exp(LeakyReLU({\overrightarrow{a}^{l}}^{T} (z_{i}^{l} || z_{j}^{l}))) }{\sum_{k \in \mathcal{N}(i)} exp(LeakyReLU({\overrightarrow{a}^{l}}^{T} (z_{i}^{l} || z_{k}^{l})))}, \\
& h_{i}^{l+1} = \sigma \left( \sum_{j \in \mathcal{N}(i)} \alpha_{ij}^{l} \left(z_{j}^{l}\right) \right), \\ \nonumber
\end{aligned}
\vspace{-1.5em}
\end{equation}
where $h_{i}^{l}$ is the lower layer embeddings and $W^{l}$ is its learnable weight matrix.
The second equation computes a pair-wise normalized attention score between neighbors, $j$ is the neighbor of node $i$. Here, it first concatenates the $z$ embeddings of the two nodes, then takes a dot product of it and a learnable weight vector $\overrightarrow{a}$, and applies a LeakyReLU function. Each node applies a softmax to normalize the attention scores on each node's incoming edges. Finally the embeddings from neighbors are aggregated together, scaled by the attention scores and applied a nonlinearity $\sigma$.

\begin{figure}
    \centering
    \includegraphics[width=0.49\textwidth]{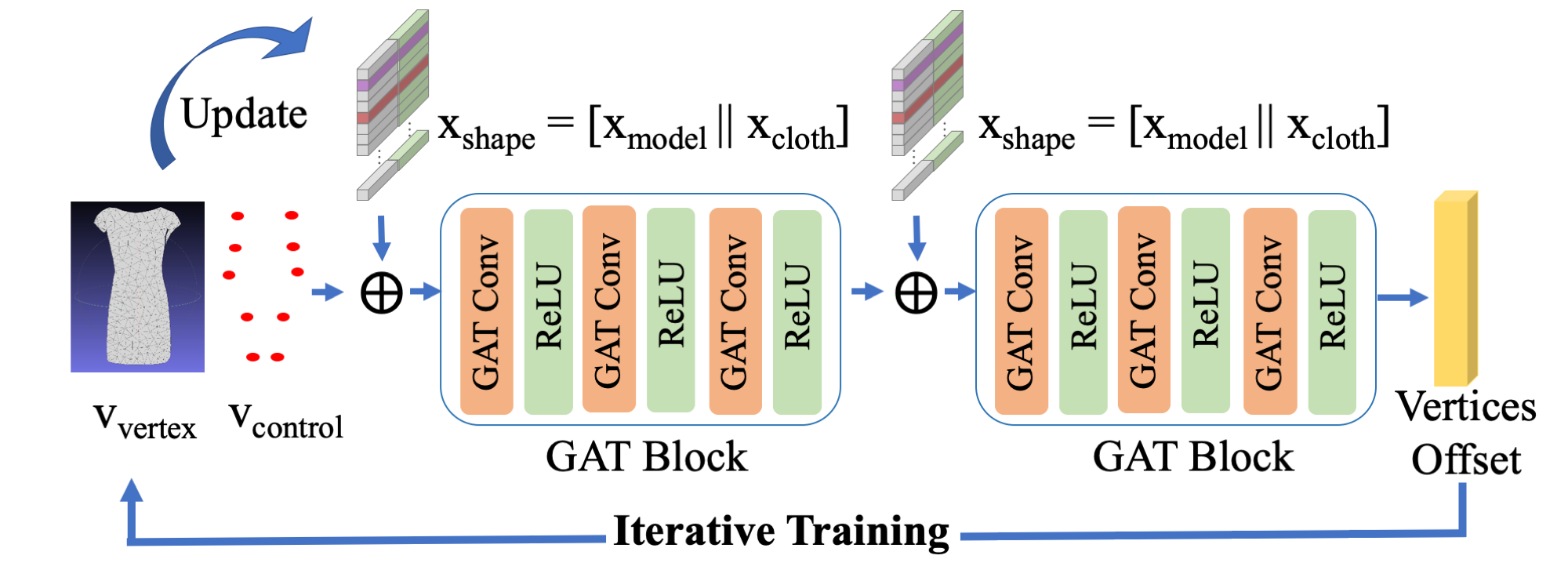}
    \caption{
    Details of the GAT based deformation module. The input contains clothing meshes and control points, as well as clothing and model features extracted by CNN extractors. The output is the vertex's coordinate offset. An iterative training strategy is adopted to optimize the result. For each iteration, model features will be recalculated according to the updated vertices coordinates.
    }
    \label{fig:4}  
\end{figure}

The network structure of adopted GAT is shown in Fig. \ref{fig:4}. Since the task is a deformation task using low-level features, shallow GAT with two graph convolution blocks are used. Each block is composed of three graph attention layers. The ground truth of GAT deformation is the offset between the vertex's coordinate in clothing image $I_c$ and generated pseudo image $I_{g'}$.
In the training stage, the vertex number $K$ of each clothing image sample could be different and can be batched using diagonal alignment. 

\subsubsection{Differentiable Rendering Module}\label{section_3_1_3}

GAT predicts coordinate offsets of all vertices, the differentiable rendering module is then used to get the warped clothing image.
We first get an indicator matrix ${\mathbb I} \in \mathbb{R} ^ {\mathbb{H} \times \mathbb{W} \times \mathbb{N}}$ during the preprocessing step, the last dimension is one-hot vector indicates which mesh each pixel belongs to ($\mathbb{N}$ is the number of meshes).
An affine transformation matrix with all meshes $M \in \mathbb{R}^{\mathbb{N} \times 2 \times 3}$ can be get by multiply the coordinates of warped mesh's vertices and the inverse of the coordinates of the original mesh's vertices.
The warped clothing image $\tilde{I_c}$ can be get by the following equation:
\begin{equation}
 \tilde{I_c} = GridSample(I_c, G_c \odot ( \mathbb{I} \times M )) ,\\ \nonumber
\end{equation}
where ${I_c}$ is the original clothing image, $G_c$ is the normalized grid of the clothing image, after getting the flow-grid of all pixels, GridSample \cite{jaderberg2015spatial} is used to get the warped clothing image $\tilde{I_c}$. This step is differentiable and can be trained in an end-to-end manner.

\subsubsection{Iterative Training Policy}\label{section_3_1_4}
In the training stage, the iterative training policy is adopted to optimize the results continuously.
After each iterative training, the Perceptual Feature Pooling is reused to extract new feature vector $x_{model}$ from the feature map $F_{model}$ according to the updated mesh coordinates, see Fig. \ref{fig:4} for more detailed information. In the experiment, SC-VTON achieves satisfying results with $2$-$3$ iterations. 

To achieve a controllable and precise result, SC-VTON should satisfy two principles under the constrain of control points.
Firstly, the deformed clothing should fit and cover the model's body as much as possible; Secondly, the deformed clothing should maintain the clothing silhouette.
The training loss $\mathcal{L}_{gat}$ of SC-VTON is composed of three parts: $\mathcal{L}_{vertex}$, $\mathcal{L}_{pixel}$ and $\mathcal{L}_{smooth}$. The first term is Huber loss \cite{huber1992robust} for constraining all vertices of the meshed clothing, the second term is the L2 pixel loss, the last term $\mathcal{L}_{smooth}$ is the smoothing constraint term for adjacent vertices:
\begin{equation}
\begin{aligned}
&\mathcal{L}_{gat}= \mathcal{L}_{vertex}+\lambda_1\mathcal{L}_{pixel}+\lambda_2\mathcal{L}_{smooth}, \\
& \mathcal{L}_{vertex}=\sum^K_{k=1}  Huber(\Delta v_k, \Delta \overline{v}_k),v_k \in V,\\
& \mathcal{L}_{pixel}= || I_c - \tilde{I_c}||^2_2, \\ \nonumber
&\mathcal{L}_{smooth}=\frac{1}{K}\sum^K_{k=1}\sum^{|Neighbor(v_k)|}_{t=1}||\Delta(v_k,v_t) - \Delta (\hat{v}_k ,\hat{v}_t)||^2_2, \\ \nonumber
\end{aligned}
\vspace{-1.5em}
\end{equation}
where $K$ denote the vertex number of all vertices $V$, $\Delta v_k$ denote the predicted vertex coordinate offset, $\Delta \overline{v}_k$ denote the ground truth of vertex coordinate offset, $|Neighbor(v_k)|$ denotes the number of the adjacent vertex set of the vertex $v_k$, $\Delta(v_k,v_t)$ denotes the vertex offset between the vertex $v_k$ and it's adjacent vertex $v_k$ in original clothing image, $\Delta(\hat{v}_k,\hat{v}_t)$ denotes the vertex offset between the vertex $\hat{v}_k$ and it's adjacent vertex $\hat{v}_k$ in the deformed clothing image, $\lambda_1,\lambda_2$ are the balance parameters.


\subsection{Self-loop Optimization with Real Pairs}\label{section_3_2}

In Section \ref{section_3_1}, large number of pseudo-labeled pairs $\{I_m, I_c, I_{g'}\}$ are adopted to train SC-VTON, the pseudo-labeled pairs are generated with an ARAP tool, which limits the performance on real data.

To solve the above drawbacks, real pairs $\{I_c, I_g\}$ of the clothing image $I_c$ and the model $I_g$ in the same clothing are adopted to optimize SC-VTON. We devise two sub-networks: a splitting network and a synthesis network, and add them into SC-VTON, which can generate self-loop supervision information for promoting robustness and performance of SC-VTON. The splitting network is devised for predicting underwear model $I'_m$ and control points from the clothed model image $I_g$. The synthesis network is devised for synthesizing the clothed model image $I'_g$ from the generated underwear model image $I'_m$ and the warped clothing image $\tilde{I_c}$ as input. The difference between real clothed model image $I_g$ and synthesis clothed image $I'_g$ will generate the supervision information for training the whole framework.

\subsubsection{Splitting Network}

As described above, the inputs of the splitting network are the real clothed model image $I_g$, the output contain the predicted underwear model image $\hat{I}_m$, underwear model's parsing result $I_{mp}$ and control points.
Before training the whole framework together, the splitting network is pre-trained by pseudo-labeled pairs $\{I_m, I_c,I_{g'}\}$ with the following loss:

\begin{equation}
\begin{split}
\mathcal{L}_{split}= ||\hat{I}_m-I_m||^2_2+
\alpha_1 \sum_{t=1}^{T}\| \phi_i(\hat{I}_m) - \phi_i(I_m) \|_1 \\ -
\alpha_2 I_{mp}log(\hat{I}_{mp})+
\alpha_3 \sum^{\hat{K}}_{k=1} ||\hat{v}^k_{control}-v^k_{control}||^2_2,\\ \nonumber
\end{split}
\vspace{-1.em}
\end{equation}
where $I_{mp}$ and $\hat{I}_{mp}$ are the origin and predicted parsing of the model,
$\hat{v}^k_{control}$ and $v^k_{control}$ are the coordinates of the predicted control points and labeled control points, 
$\phi_i(I_m)$ and $\phi_i(\hat{I}_m)$ denote the feature maps of image $I_m$ and generated image $\hat{I}_m$ of the $t$-th layer ($T = 4$) in the visual perception network $\phi$, which is a VGG19 \cite{simonyan2014very} pre-trained on ImageNet \cite{deng2009imagenet}. 
$\alpha_1,\alpha_2,\alpha_3$ are the balance parameters.

\subsubsection{Synthesis Network}

As depicted in Fig. \ref{fig:2}, the synthesis network contains three input parts: the warped clothing image, the underwear model image, and the corresponding parsing. 
The output of the synthesis network is the synthesis clothed model image. Similarly, the synthesis network are also pre-trained on the pseudo-labeled pairs $\{I_m, I_c,I_{g'}\}$ with the following loss:
\begin{equation}
\mathcal{L}_{synth}=  ||\hat{I}_{g'}-I_{g'}||^2_2 + \beta \sum_{t=1}^{T}\| \phi_i(\hat{I}_{g'}) - \phi_i(I_{g'}) \|_1,\\ \nonumber
\end{equation}
where $\hat{I}_{g'}$ is the synthetic clothed model image with the pseudo-labeled pair as input. $\beta$ is the balance parameter.

\subsubsection{Whole Framework Optimization}

With the pre-trained splitting network and the synthesis network, the whole framework can be trained on real pairs $\{I_c, I_g\}$ with the following loss:
\begin{equation}
\mathcal{L}_{real}= ||\hat{I}_g-I_{g}||^2_ 2+ \gamma \sum_{t=1}^{T}\| \phi_i(\hat{I}_g) - \phi_i(I_g) \|_1,\\ \nonumber
\end{equation}
where $\hat{I}_g$ is the synthetic clothed model image of whole framework, $I_g$ is the input with the real clothed model image, $\gamma$ is the balance parameter.

In the training stage of the whole framework, SC-VTON is also trained on the pseudo-labeled pairs simultaneously.
The total loss $\mathcal{L}$ of the whole framework is defined as follows:
\begin{equation}
\mathcal{L}=\mathcal{L}_{synth} + \eta \mathcal{L}_{real},\\ \nonumber
\end{equation}
where $\eta$ is the balance parameter.

\subsubsection{Extension to Clothed Model Virtual Try-on Task}
With the splitting network and the synthesis network, the whole framework can be easily extended to the virtual try-on task for the clothed model image. 
For a clothed model image, the splitting network first predicts the underwear model image, parsing, and control points. Then GAT uses the meshed clothing, predicted underwear model, and control points as input to get the warped clothing image. The synthesis network synthesizes the final result.

\section{Experiments}

\subsection{Implementation Details}\label{section_4_1}

\textbf{Underwear Model Virtual Try-On (UMV) Dataset.}
The UVM dataset consists of $143,428$ pseudo labeled pairs in total. To get the pseudo labeled pairs, we photographed $2648$ underwear models with different poses and collected tens of thousands of clothing images from the e-commerce platform. Key points and parsing labels for both models and clothing are defined.
We manually label the model images to get accurate results. For clothing images, we use pre-trained models \cite{li2019self,sun2019high} to predict key points and parsing results, and remap to our own definitions.
An ARAP based annotation tool is used with manually fine-tune to get the pseudo label $I_{g'}$.
\emph{supplementary materials} provide more detailed descriptions of the UMV dataset.

\textbf{ Multi-Pose Virtual Try-On (MPV) Dataset.}
MPV dataset \cite{dong2019towards} consists of $37,723$/$14,360$ person/clothing images, the dataset used for experiments is one part of the MPV dataset, consisting of 14,754 pairs of top clothing images and positive perspective images of female models. The resolution of images in the dataset is $256 \times 192$.

\textbf{Network Architecture.}
The GAT module contains two graph attention blocks, each with three graph attention layers followed by a ReLU activation layer. The output dimension of each layer is $256$, and the final output dimension is $2$. 
The feature maps of the former three layers with $64$, $128$, $256$ channels of VGG network are used in the shape information extraction module.
For each vertex, Perceptual Feature Pooling extract the feature from four near pixels using bilinear interpolation according to its coordinate position and get a $1\times 448$ vector.
Two $1\times 448$ vectors extracted from the model and the clothing are concatenated into a $1\times 896$ vector, and then reduce the dimension by a fully connection layer to get 
the final $1\times 256$ vector.

The Splitting Network is an U-Net \cite{ronneberger2015u} like encoder-decoder network. Three contracting blocks are used in Encoder, each with two convolution layers, the output stride is $8$. The decoder uses three expansive blocks, each with two convolution layers and a transposed convolution to recover feature map size and concatenate with the corresponding contracting block's output. The Synthesis Network's has the similar network architecture as the Splitting Network. More details about network architecture are given in the \emph{supplementary materials}.

\textbf{Parameters Setting.} 
As for training parameters, we set batch size $4$, the epoch is $30$, and $2$ iterations are used. ADAM optimizer is used with $\alpha = 0.5$ and $\beta = 0.999$, the learning rate is $10^{-3}$. We set 
$\lambda_1 = 1$, 
$\lambda_2 = 10$,
$\alpha_1 = 0.1$, 
$\alpha_2 = 1$,
$\alpha_3 = 1$, 
$\beta = 0.1$, 
$\gamma = 0.1$, 
$\eta = 1$.
Hyperparameters are set in terms of the importance of each subterm, and the balance of magnitudes between each loss.

\begin{figure}
    \centering
    \includegraphics[width=0.49\textwidth]{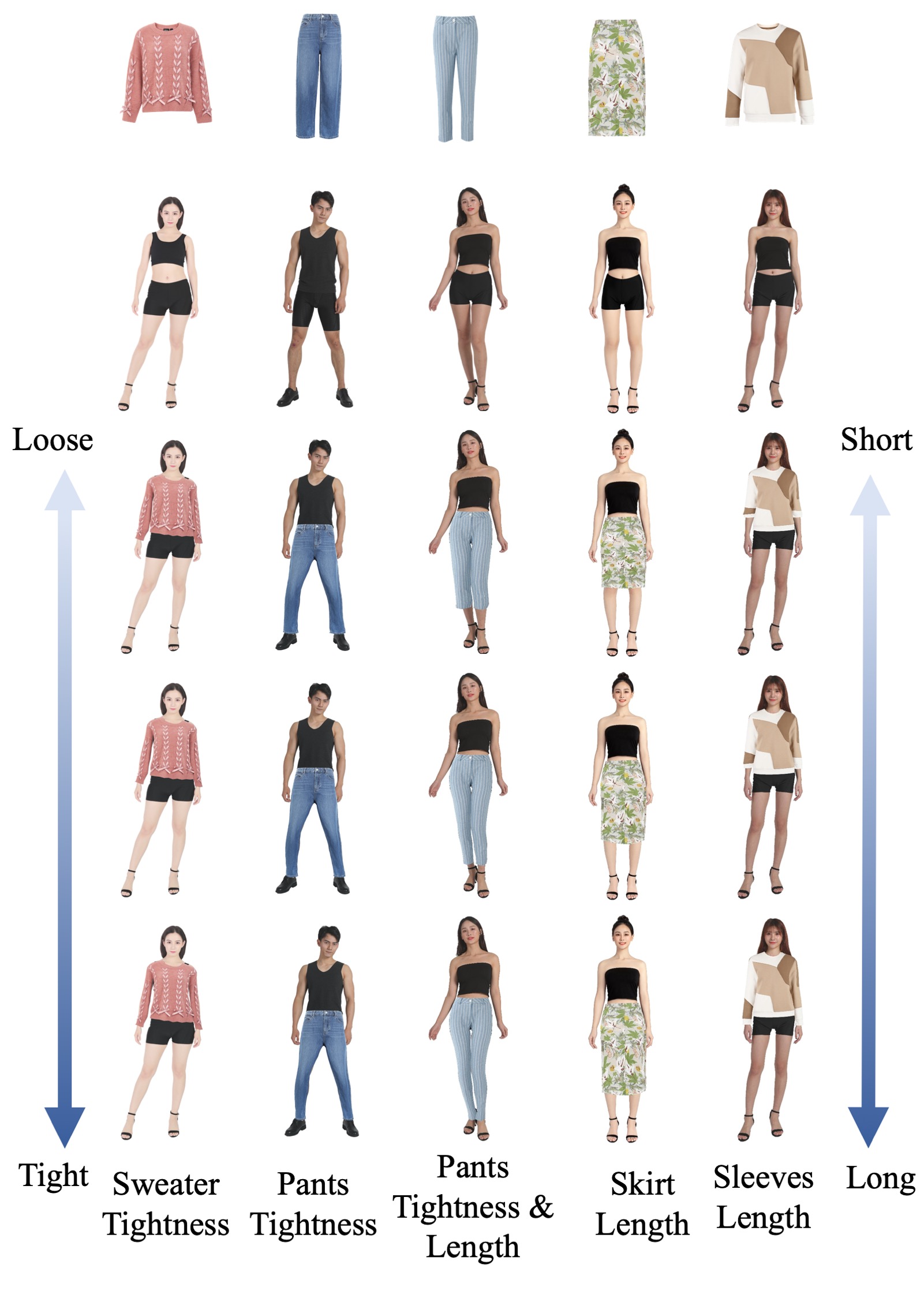}
    \caption{
    Controllable shape results with different control point setting. The shape can be continuously controlled in length and tightness.
    }
    \label{fig:5}
\end{figure}

\begin{figure*}
    \centering
    \includegraphics[width=0.9\textwidth]{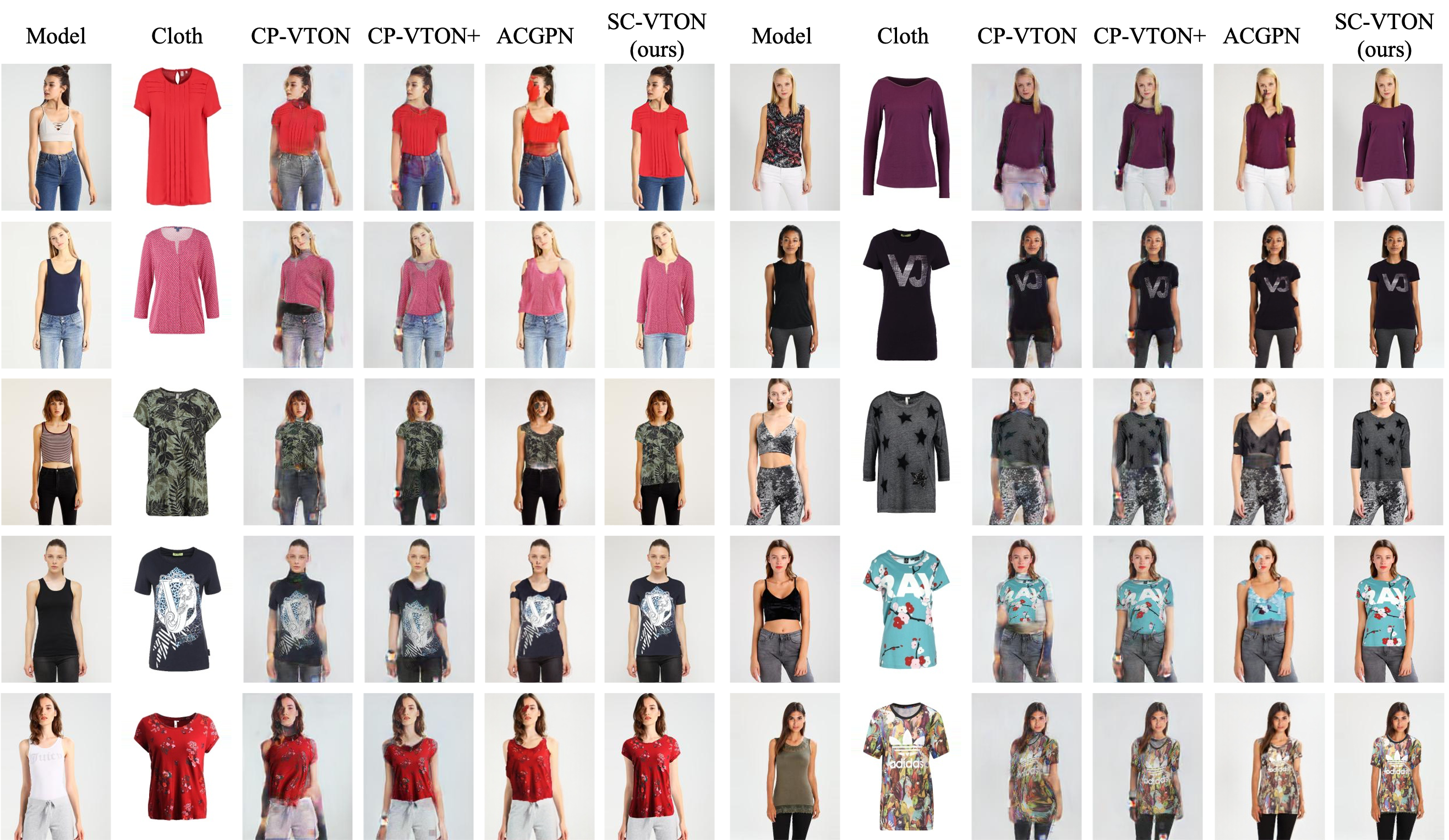}
    \caption{Visual comparison with CP-VTON \cite{wang2018toward}, CP-VTON+ \cite{Minar_CPP_2020_CVPR_Workshops} and ACGPN \cite{yang2020towards}. The textures of clothing range from simple to complex, our method is better at keeping the clothing's texture and maintaining the clothing's silhouette.}
    \label{fig:6}
\end{figure*}

\subsection{Experiment on Controllable Shape}\label{section_4_2}
To verify the effectiveness of shape control, we set different control points as guidance for the same clothing and underwear model. Fig. \ref{fig:5} shows the results of various models and clothing, where we can see that the tightness of sweater and pants, the length of skirt and sleeves are successfully controlled with different control points as inputs. What's more, for the pants in the middle column of Fig. \ref{fig:5}, the tightness and length are controlled simultaneously.

\subsection{Qualitative Results}\label{section_4_3}

Fig. \ref{fig:6} illustrates our results compared with other methods on test samples selected from the MPV dataset. CP-VTON \cite{wang2018toward}, CP-VTON+ \cite{Minar_CPP_2020_CVPR_Workshops} and ACGPN \cite{yang2020towards} are adopted for comparison. The textures of clothing range from simple to complex.
Compared with CP-VTON and CP-VTON+, the texture of our approach is much clearer, please see the results of the last two rows in Fig. \ref{fig:6}. Meanwhile, compared with ACGPN, our approach can better maintain the clothing's silhouette. For example, for the two cases in the last row in Fig. \ref{fig:6}, ACGPN has turned short sleeves into sleeveless, which is an obvious mistake.
Overall, our method is better at keeping the clothing's texture and can better maintain the clothing's silhouette. 

\subsection{Quantitative Results}\label{section_4_4}
We conduct a user study on the Qince Platform (A crowdsourcing platform developed by Alibaba Inc.)
Given the clothing image, and two synthesized clothed model images by two different methods (with resolution $256 \times 192$ and $512 \times 384$, respectively), the worker is asked to choose the one is more realistic and accurate in a virtual try-on situation. 
We choose $500$ testing samples from the MPV dataset.
All synthesized results are presented five times to different workers to avoid individual preferences. Quantitative comparisons are summarized in Table \ref{table_1}. As can see from the table, our method gives better results than the other methods at a resolution of $ 256 \times 192 $. When the resolution is enlarged to $512 \times 384$, our evaluation results are significantly better than the other methods, demonstrating that our method can better maintain texture details for large resolution images.

\begin{table}
\centering
\caption{ Perceptual user study results with other methods on the MPV dataset.
The score represents the proportion of samples that workers considered more realistic.}
\begin{tabular}{l|c|c}
\hline
Method   & Resolution 256 & Resolution 512 \\ \hline \hline
CP-VTON \cite{wang2018toward}  & 0.37            & 0.27            \\ 
SC-VTON (Ours)  & 0.63            & 0.73            \\ \hline
CP-VTON+ \cite{Minar_CPP_2020_CVPR_Workshops} & 0.43            & 0.33            \\ 
SC-VTON (Ours)  & 0.57            & 0.67            \\ \hline
ACGPN \cite{yang2020towards}    & 0.42            & 0.38            \\ 
SC-VTON (Ours)  & 0.58            & 0.62            \\ \hline
\end{tabular}
\label{table_1}
\end{table}

\subsection{Comparing the Warping Results}\label{section_4_5}

\begin{figure}[!t]
    \centering
    \includegraphics[width=0.4\textwidth]{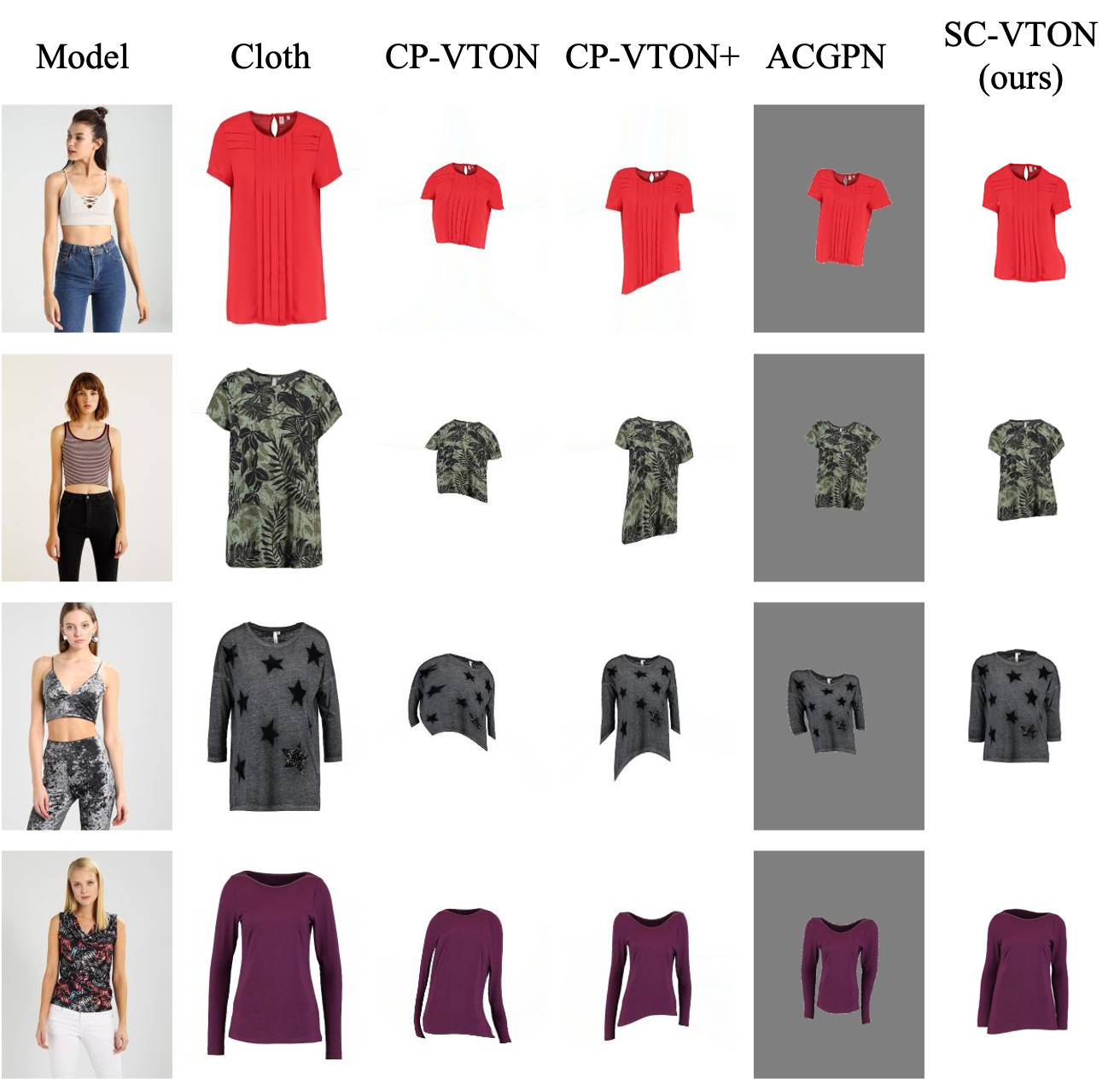}
    \caption{Warping results with CP-VTON, CP-VTON+ and ACGPN. Compared with other methods, our warping results are flatter and rigid, and also better maintain the original silhouette of the garment.}
    \label{fig:7}
\end{figure}

\begin{table}[!t]
\centering
\caption{ Precision within a certain threshold on upper clothing, pants and their average. Larger is better. SC-VTON$^{\dagger}$ is the optimal model trained with UMV. SC-VTON$^{\ast}$ is the optimal model trained with both UMV and MPV.}
\begin{tabular}{l|c|c|c}
\hline
Method & Uppers & Pants & Average\\ \hline \hline
SC-VTON (w/o cloth info)                         & 0.67 & 0.80 & 0.74  \\ \hline
SC-VTON (w/o $\mathcal{L}_{smooth}$)              & 0.84 & 0.87 & 0.86     \\ \hline
SC-VTON (w/o iterative training)                           & 0.71 & 0.83 & 0.77     \\ \hline
SC-VTON$^{\dagger}$ (UMV)                             & 0.76 & 0.83 & 0.80     \\ \hline
SC-VTON$^{\ast}$ (UMV \& MPV)                                & 0.78 & 0.85 & 0.82     \\ \hline
\end{tabular}
\label{table_2}
\end{table}

As shown in Fig. \ref{fig:7}, we further analyze and show the visual comparison of warping modules with CP-VTON, CP-VTON+ and ACGPN on MPV dataset.
Both CP-VTON and CP-VTON+ use the Spatial Transformation Network (STN) to learn parameters of Thin-Plate Spline (TPS) to warp the clothing image.
ACGPN uses TPS with a second-order difference constraint to get less distortion results.
As shown in Fig. \ref{fig:7}, the results of CP-VTON and CP-VTON+ are often distorted in the sleeve and torso parts, and ACGPN is prone to skewing.
This is because the deformation between different clothing parts can affect each other using TPS.

We take this into account in the pre-processing stage. For the parsing definitions of person and clothing, we distinguish between different parts. For example, the left arm, torso and right arm are marked as different parsing categories (See \emph{supplementary materials}), thus we know which part each mesh belongs to when meshing the clothing.
GAT allows overlap between different meshes, and different parts do not affect each other. During the rendering step, the overlapping regions can be handled orderly. 
Therefore, compared to other methods, our warping results are flatter and rigid, and also better maintain the original silhouette of the garment.



\subsection{Ablation Experiments}\label{section_4_6}

Here we conducted a series of ablation experiments to verify each module and loss's effectiveness of the proposed methods. For evaluation metric, we calculate the precision by checking the percentage of vertices between prediction and ground truth within a certain threshold $\tau=10^{-2}$.

\begin{figure}[!t]
    \centering
    \includegraphics[width=0.49\textwidth]{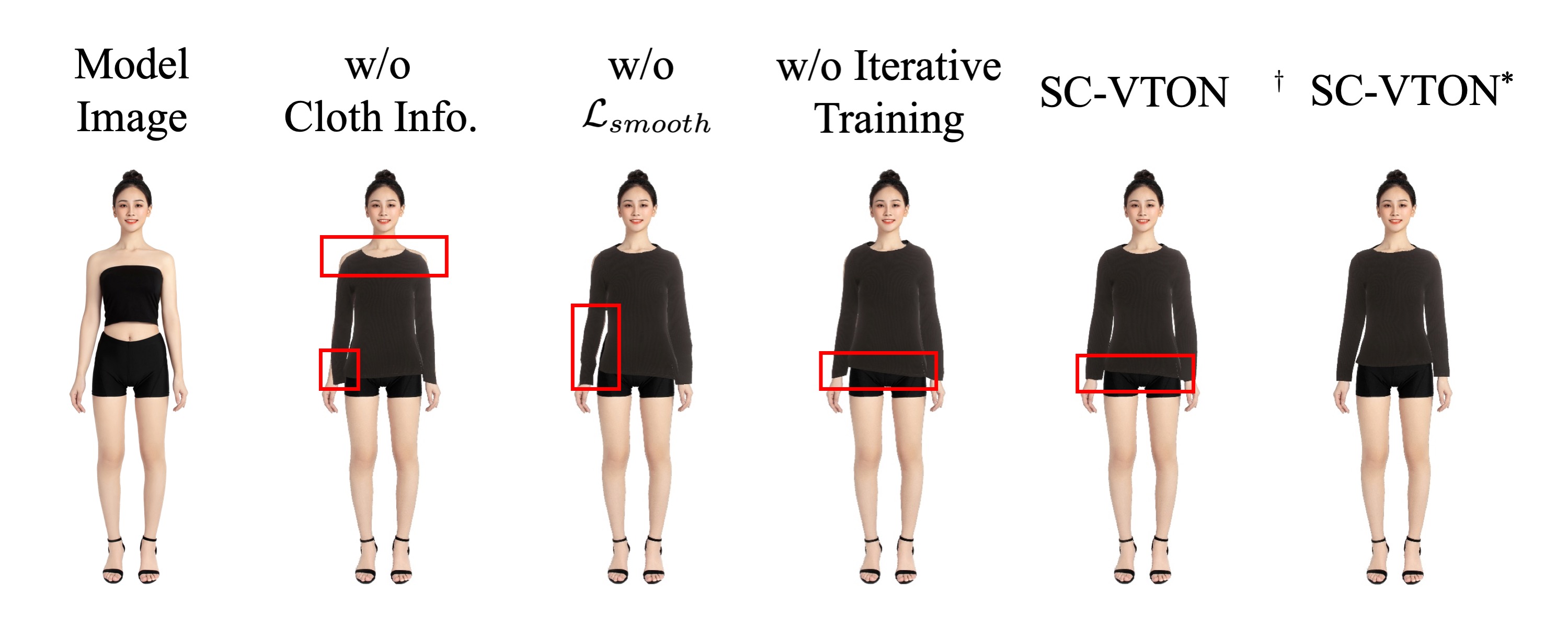}
    \caption{Visual results of ablation study experiments. From left to right: model image, result without using clothing infomation in GAT, result without using $\mathcal{L}_{smooth}$, result without using iterative training in GAT. SC-VTON$^{\dagger}$ is the optimal model trained with UMV. SC-VTON$^{\ast}$ is the optimal model trained with both UMV and MPV.}
    \label{fig:8}
\end{figure}

Table \ref{table_2} and Fig. \ref{fig:8} show the quantitative and qualitative results of the ablation study. SC-VTON$^{\dagger}$ is the optimal model trained with UMV dataset. SC-VTON$^{\ast}$ is the optimal model trained with both UMV and MPV datasets. From Table \ref{table_2} and Fig. \ref{fig:8}, we can see that removing clothing's image and parsing information made the results worse.
Removing $\mathcal{L}_{smooth}$ increases the accuracy, but the silhouette of the clothing becomes jagged.
There is also a slight decrease with no iteration during training, demonstrating that iterative training can improve the deformation result.
Training with UMV and MPV simultaneously also slightly boosts the performance.

\subsection{Extend to Typical Virtual Try-on Task}\label{section_4_7}
As described in Section \ref{section_3_2}, with the splitting network and the synthesis network, the whole framework can be extended to typical virtual try-on task for the clothed model. 
At inference time, both the clothed model image $I_m$ and a different clothing image $I_{c'}$ are sent to the network. We can directly put the warped clothing image to the predicted underwear model to get the final result, or use the Synthesis Network to generate a refined result. 
More results are given in the \emph{supplementary materials}.
The task of predicting the underwear model is very difficult, hence the visual result of \uppercase\expandafter{\romannumeral2} is not quite good. Even so, adding \uppercase\expandafter{\romannumeral2} can optimize the warping result, while increasing the robustness on new samples.


\subsection{Failure Cases}\label{section_4_8}
\begin{figure}
    \centering
    \includegraphics[width=0.49\textwidth]{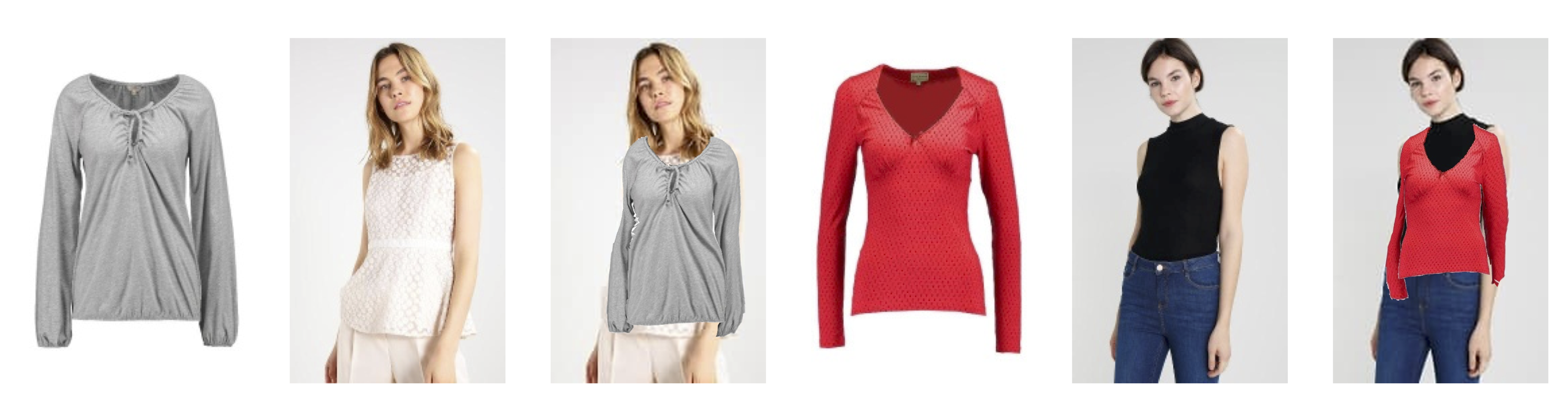}
    \caption{Failure cases. Our method fails on side posture with large angles. For the obscured part (left arm), the result of deformation is wrong.}
    \label{fig:9}
\end{figure}

All warping methods based on 2D clothing images face a common problem: cannot handle the side postures. Most works infer the side textures of the clothing by generators, the results are often blurry and not realistic. Our method also has this problem and this paper only deal with the front view models. For side posture cases, the results are not good, please see Fig. \ref{fig:9} for visual results.
We came up with some solutions to solve the problem as our future works: for person modeling, 3D information need to be considered, DensePose \cite{guler2018densepose} or SMPL \cite{loper2015smpl} model may be two options. A more detailed clothing dataset need to be collected, which should contain at least two clothing images of front and back views. For the warping algorithm, depth information need to be considered during the warping process.


\section{Conclusion}\label{section_5}

In this paper, we propose a Shape Controllable Virtual Try-On Network (SC-VTON) to wear clothing for underwear models, which is an urgent demand for an online clothing shop to exhibit new clothing efficiently. The proposed method comprises two parts: GAT based clothing deformation for the underwear model and self-loop optimization with real pairs. The former is devised for deforming clothing for the underwear model under the constraint of shape control points. The latter is introduced for improving the robustness and performance of SC-VTON, which also extends the whole framework to the typical virtual try-on task.
The advantage of the proposed method is that SC-VTON can achieve continuous shape control in length and tightness. 
As far as we know, SC-VTON is the first shape controllable method for the 2D image-based virtual try-on task.
What's more, based on the inherent advantage of the underwear model image, we can extract more accurate model's figure information than the clothed model, which brings in more pleasing virtual try-on visual results.
Our method can maintain detailed texture and clothing silhouette for high-resolution image, which has practical advantages in real applications.

%
\begin{acks}
This work is supported by Key Research and Development Program of Zhejiang Province (2018C01004), National Natural Science Foundation of China (61976186,U20B2066), Fundamental Research Funds for the Central Universities (2021FZZX001-23), Alibaba-Zhejiang University Joint Research Institute of Frontier Technologies.  
\end{acks}


\bibliographystyle{ACM-Reference-Format}
\bibliography{ms}

\appendix









\end{document}